\documentclass{article}
\usepackage[utf8]{inputenc}
\usepackage[]{graphicx}
\usepackage{titling}
\usepackage[margin=35mm]{geometry}
\usepackage{url}     
\usepackage{hyperref}
\usepackage{dirtree}
\usepackage{subfig}
\hypersetup{
    colorlinks=true,
    linkcolor=blue,
    filecolor=magenta,      
    urlcolor=cyan,
}

\newcommand{\etal}{\emph{et~al.}}

\pretitle{\centering\Huge\bfseries}
\posttitle{\par\vskip 0.5em}
\preauthor{\Large}
\postauthor{}
\predate{\par\large\centering}
\postdate{\par}

\title{DeepSeagrass Dataset}
\author{ \begin{center} Scarlett Raine, Ross Marchant, Peyman Moghadam, Frederic Maire, Brett Kettle and Brano Kusy \end{center}  }
\date{}

\renewenvironment{abstract}
 {\small
  \begin{center}
  \bfseries \abstractname\vspace{-.5em}\vspace{0pt}
  \end{center}
  \list{}{
    \setlength{\leftmargin}{.5cm}%
    \setlength{\rightmargin}{\leftmargin}%
  }%
  \item\relax}
 {\endlist}

\begin{document}

\maketitle

{\centering \large CSIRO Data61 and \\ QUT Centre for Robotics, Brisbane, Australia \par}
\vspace{0.5cm}

\begin{abstract}
\noindent We introduce a dataset of seagrass images collected by a biologist snorkelling in Moreton Bay, Queensland, Australia.  The images are labelled at the image-level by collecting images of the same morphotype in a folder hierarchy. We also release pre-trained models and training codes for detection and classification of seagrass species at the patch level at: \url{https://github.com/csiro-robotics/deepseagrass}. 
\end{abstract}

\section{Introduction}
Seagrasses provide significant value in terms of ecosystem services, such as protection of coastlines, improvement of water quality, increasing productivity of fisheries and acting as `blue carbon' sinks \cite{lavery}.  Management of seagrass beds requires broad scale accurate surveys \cite{unsworth}.  Recent approaches to broad-scale surveys include remote sensing (satellites, conventional aerial photography, drones) and underwater vehicles that are towed (TUVs \cite{rende}), remotely operated (ROVs \cite{finkl}) or fully autonomous (AUVs \cite{monk}). Computer vision and machine learning techniques are required to automatically and efficiently analyse the large amount of image data produced using these methods \cite{coralnet}. Prior approaches to seagrass detection and classification have not offered a solution for the multi-species case.  This work introduces a multi-species seagrass dataset of image patches.  We release the training and validation datasets but retain the test dataset.  We also release the pre-trained models and code used for training and evaluation at \url{https://github.com/csiro-robotics/deepseagrass}.  The dataset is available for download from the CSIRO data portal at \url{https://doi.org/10.25919/spmy-5151}.  When making use of this dataset we ask that Raine~\etal{}~\cite{raine2020multispecies} is cited. 

\section{The Dataset}
This section introduces and describes the DeepSeagrass dataset. Raine~\etal{}~\cite{raine2020multispecies} outlines the details of the multi-species detector and classifier for seagrasses based on a deep convolutional neural network. 
The data collection methodology is described in Section \ref{subsect:obtain_data}.  The data was prepared for training a deep learning architecture following the process outlined in Section \ref{subsect:process}.  Finally, a number of example images are presented in Section \ref{subsect:vis}.

\subsection{Obtaining the Dataset}
\label{subsect:obtain_data}
Images were acquired across nine different seagrass beds in Moreton Bay, Australia over four days during February 2020. Search locations were chosen according to distributions reported in the publicly available dataset \cite{roelfsema}. A biologist made a search of each area, snorkelling in approximately 1 - 2m of water during low to mid tide. In-situ search of seagrass beds resulted in batches of photographs in 78 distinct geographic sub-areas, each containing one particular seagrass morphotype (or bare substrate).  Images were taken using a Sony Action Cam FDR-3000X from approximately 0.5m off the seafloor at an oblique angle of around 45 degrees. Over 12,000 high-resolution (4624 x 2600 pixels) images were obtained. 

Images were reviewed to ensure that any containing a second morphotype at more than approximately 0.5\% density were placed into a `mixed' class. The remainder were labelled according to their dominant morphotype, and divided into three categories (dense, medium, sparse) according to the relative density of seagrass present, for a total of 11 classes.

\begin{figure}
    \centering
    \includegraphics[width=0.45\textwidth]{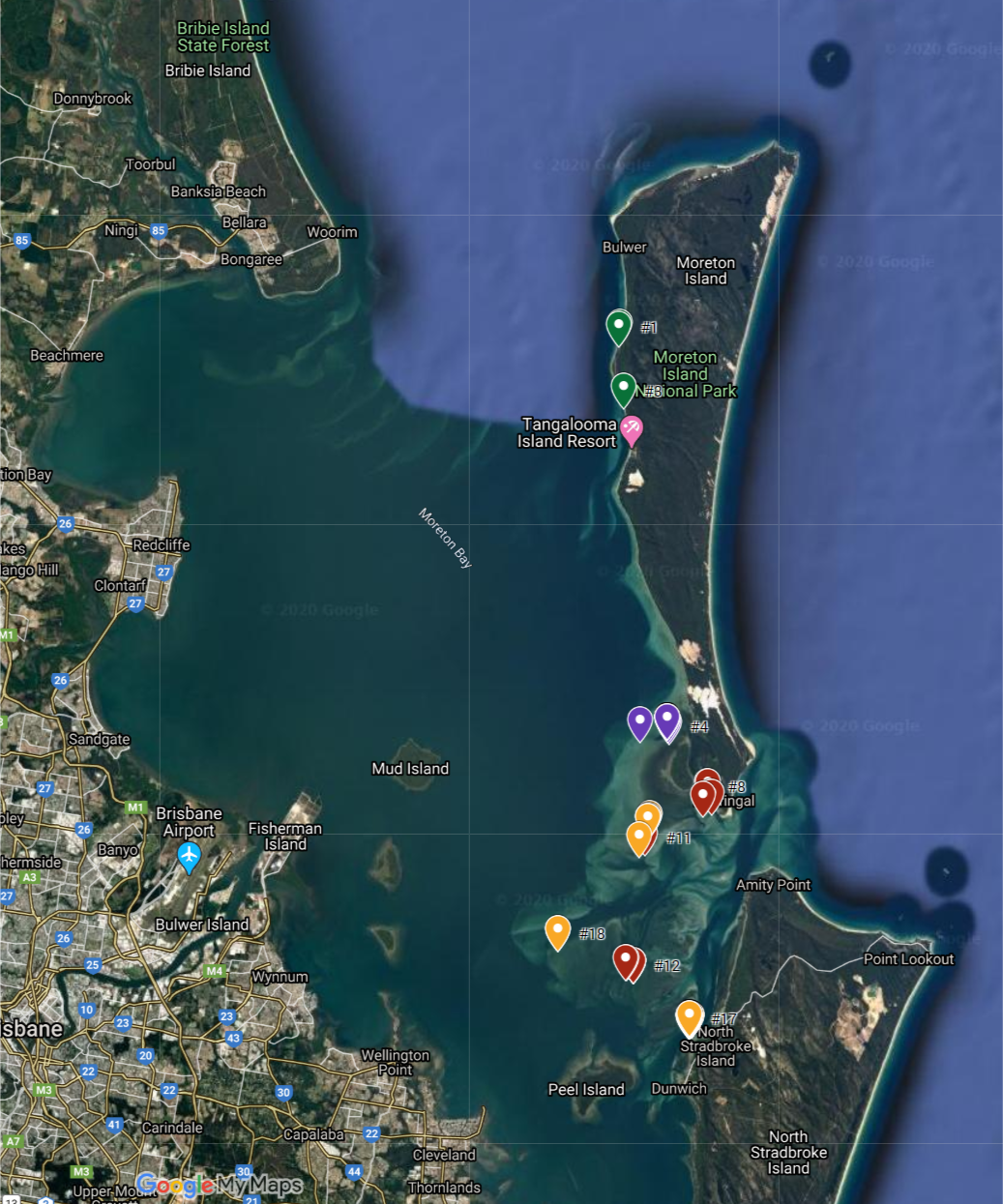}
    \caption{Map of 78 distinct sub-areas for our \textit{DeepSeagrass} dataset. Colour of the marker refers to the date of collection, green: 2020-02-07, purple: 2020-02-08, red: 2020-02-10, orange: 2020-02-11.}
    \label{fig:map}
\end{figure}

\subsection{Dataset Processing}
\label{subsect:process}
The dataset was then prepared for use in a machine learning pipeline by taking the dense seagrass images and dividing them into train and test sets. Images from 70 of the sub-areas were allocated to the training set (1,701 images in total), while images from the remaining eight sub-areas were reserved exclusively for the test set (335 images). Different sub-areas were used so that evaluation of a classification system on the test set would assess how well it generalised to different geographic areas.

The dataset is comprised of three seagrass morphotypes and a class for images not containing seagrass. Together with the test set patches, this yields a total of 66,946 patches (Table \ref{table:class_dist}).

\begin{table}
\begin{center}
\begin{tabular}{ |c|c|c|c|c| }
    \hline 
        \multicolumn{5}{|c|}{\textbf{Training}}\\
    \hline
    \textbf{Strappy} & \textbf{Ferny} & \textbf{Rounded} & \textbf{Background} & \textbf{Total} \\ 
    \hline
    11,584 & 8,256 & 13,792 & 9,216 & 42,848 \\ 
    \hline
    \multicolumn{5}{|c|}{\textbf{Validation}}\\
    \hline
    \textbf{Strappy} & \textbf{Ferny} & \textbf{Rounded} & \textbf{Background} & \textbf{Total}\\ 
    \hline
    2,880 & 2,080 & 3,456 & 2,304 & 10,720 \\
    \hline 
        \multicolumn{5}{|c|}{\textbf{Test}}\\
    \hline
    \textbf{Strappy} & \textbf{Ferny} & \textbf{Rounded} & \textbf{Background} & \textbf{Total} \\ 
    \hline
    2,643 & 4,447 & 1,345 & 4,943 & 13,378 \\
    \hline 
\end{tabular}
\end{center}
\caption{Distribution of image patches in DeepSeagrass dataset}
\label{table:class_dist}
\end{table} 

 We use underwater images that consist of only one seagrass morphotype in each image. For the \textit{DeepSeagrass} dataset, these are the `Ferny - dense', `Strappy - dense' and `Round - dense' classes, plus images with no seagrass from the `Substrate' class.
 
\begin{figure}[h]
\centering
\includegraphics[scale=0.6]{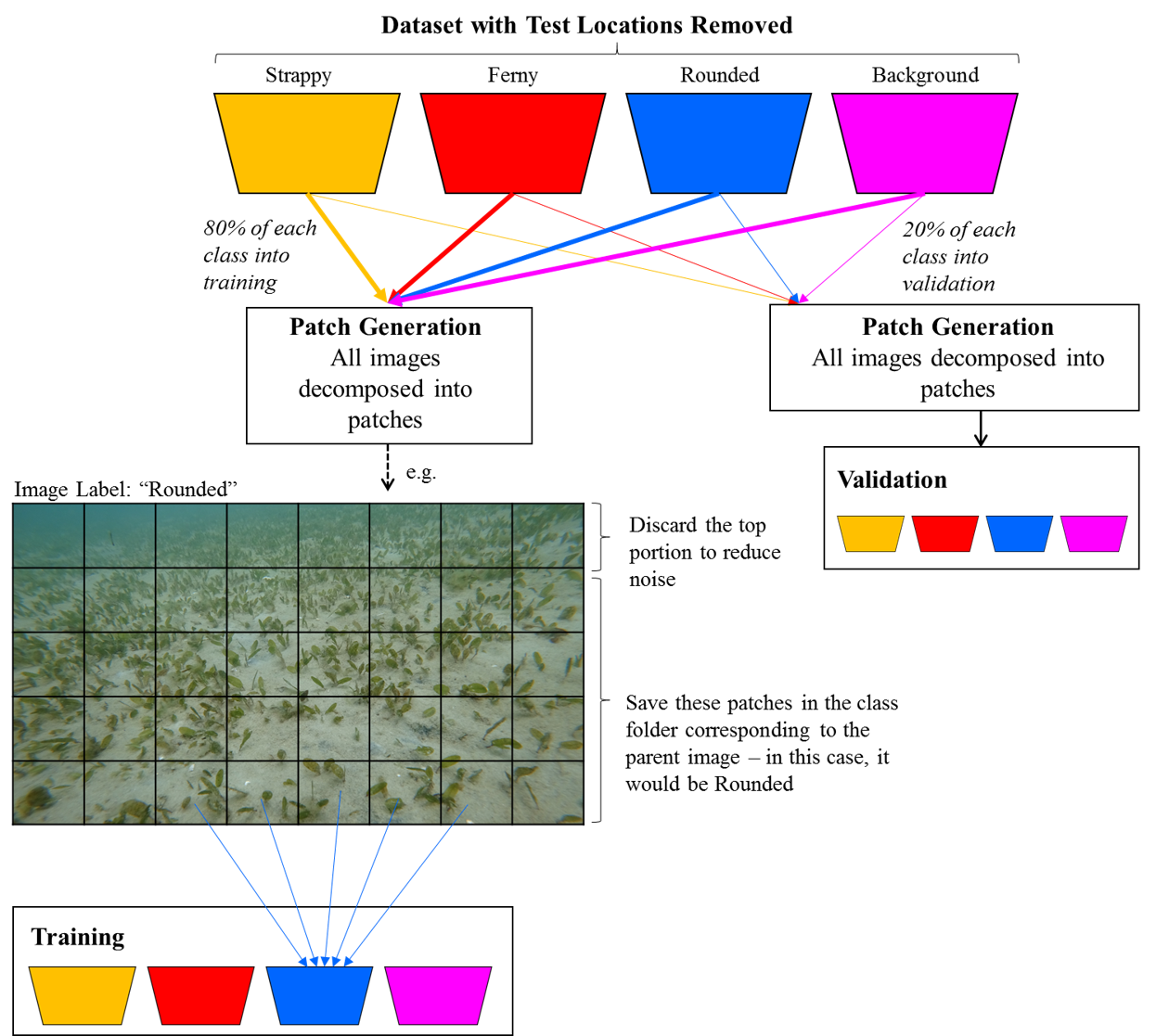}
\caption{Generating the static dataset}
\label{fig:data_gen}
\end{figure}

 Images are then divided into a grid of patches. For the \textit{DeepSeagrass} dataset we used a grid of 5 rows by 8 columns to generate 40 patches of size 520x578 pixels per image. The top row of patches for each image was discarded, as their oblique pose in low underwater visibility conditions frequently resulted in hard to distinguish seagrass (Figure \ref{fig:data_gen}). 
 
The file structure following patch generation
is as follows:\newline
\dirtree{%
.1 training.
.2 Strappy.
.3 Image0\_Row1\_Col0.jpg.
.3 Image0\_Row1\_Col1.jpg etc.
.2 Ferny.
.3 Image0\_Row1\_Col0.jpg.
.2 Rounded.
.3 Image0\_Row1\_Col0.jpg.
.2 Background.
.3 Image0\_Row1\_Col0.jpg.
.1 validate.
.2 Strappy.
.3 Image0\_Row1\_Col0.jpg.
.2 Ferny.
.3 Image0\_Row1\_Col0.jpg.
.2 Rounded.
.3 Image0\_Row1\_Col0.jpg.
.2 Background.
.3 Image0\_Row1\_Col0.jpg.
.1 test.
.2 Strappy.
.3 Image0\_Row1\_Col0.jpg.
.2 Ferny.
.3 Image0\_Row1\_Col0.jpg.
.2 Rounded.
.3 Image0\_Row1\_Col0.jpg.
.2 Background.
.3 Image0\_Row1\_Col0.jpg.
}


Under this naming convention, patches are labelled with their parent image and the row and column they originated from.  Note that Row 0 was not used in the DeepSeagrass dataset as generally this row of images contained indistinguishable seagrass.

The training and validation dataset consists of 10 folders, with 53,568 seagrass patches to total 4.12GB of data.   The separate test dataset contains 13,378 image patches to a total of 1.05GB.  The dataset is available for download from the CSIRO data portal at: \url{https://doi.org/10.25919/spmy-5151}.  The dataset contains necessary information and images for the 5-class classifier discussed in our paper, however it is not available at the 289 x 260 patch size.

\subsubsection{5-Class Case}
\label{5-class}
Our paper discussed an extension of the approach to a 5-class detector and classifier.  This method involved separation of the `Background' class into `Substrate' and `Water' column classes.  If using this version of the dataset, the additional folders contained within the `For 5-Class Case' folder must be added. For the training and validation datasets, this requires performing a split on the `Water' folder and adding it to the existing folders.  The existing `Background' folder contains predominantly substrate images in this case.  For the test dataset, the `Background' folder must be replaced with the provided `Substrate' and `Water' folders. 

The file structure following patch generation for the five class case is as follows: \newline
\linebreak
\dirtree{%
.1 training.
.2 Strappy.
.3 Image0\_Row1\_Col0.jpg.
.3 Image0\_Row1\_Col1.jpg.
.3 Image0\_Row1\_Col2.jpg etc.
.2 Ferny.
.3 Image0\_Row1\_Col0.jpg.
.2 Rounded.
.3 Image0\_Row1\_Col0.jpg.
.2 Background.
.3 Image0\_Row1\_Col0.jpg.
.2 Water.
.3 Image0\_Row1\_Col0.jpg.
.1 validate.
.2 Strappy.
.3 Image0\_Row1\_Col0.jpg.
.2 Ferny.
.3 Image0\_Row1\_Col0.jpg.
.2 Rounded.
.3 Image0\_Row1\_Col0.jpg.
.2 Background.
.3 Image0\_Row1\_Col0.jpg.
.2 Water.
.3 Image0\_Row1\_Col0.jpg.
.1 test.
.2 Strappy.
.3 Image0\_Row1\_Col0.jpg.
.2 Ferny.
.3 Image0\_Row1\_Col0.jpg.
.2 Rounded.
.3 Image0\_Row1\_Col0.jpg.
.2 Substrate.
.3 Image0\_Row1\_Col0.jpg.
.2 Water.
.3 Image0\_Row1\_Col0.jpg.
}

\subsection{Example Dataset Visualisations}
\label{subsect:vis}
\begin{figure}
\centering
\begin{tabular}{cccc}
    \includegraphics[width=35mm]{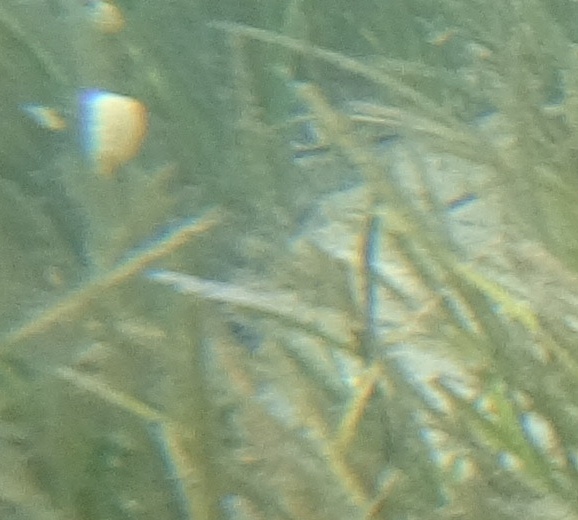} &
    \includegraphics[width=35mm]{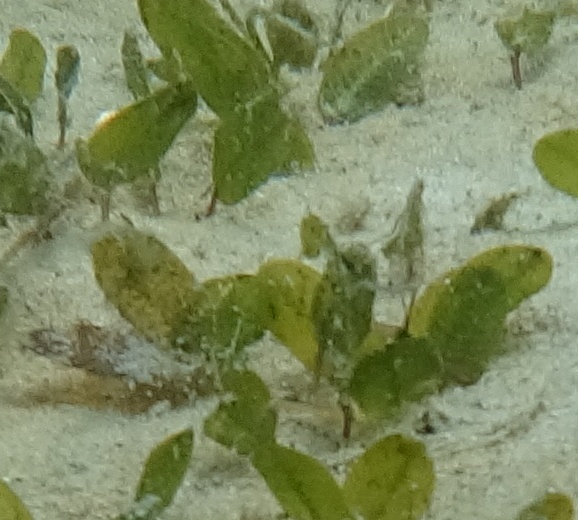} &
    \includegraphics[width=35mm]{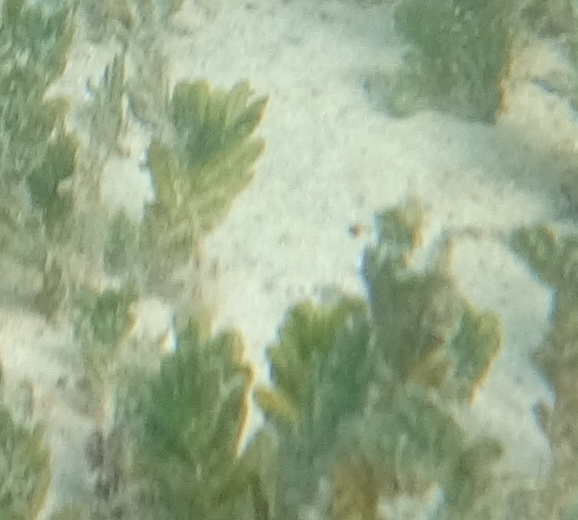} & 
    \includegraphics[width=35mm]{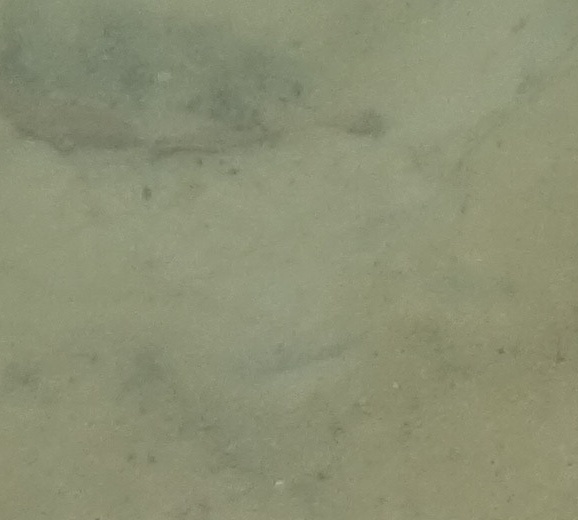} 
    \\
    \includegraphics[width=35mm]{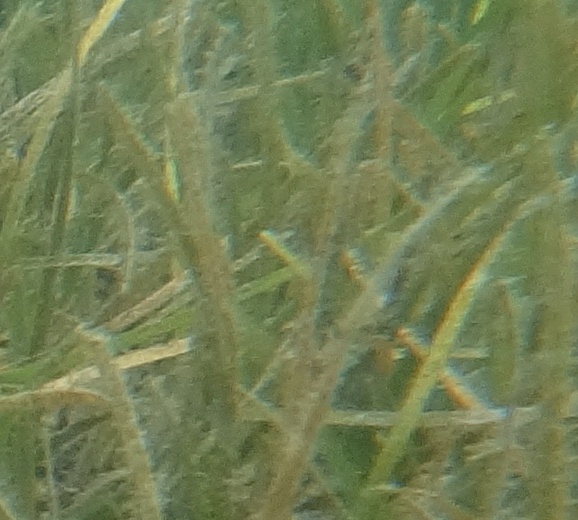} &  
    \includegraphics[width=35mm]{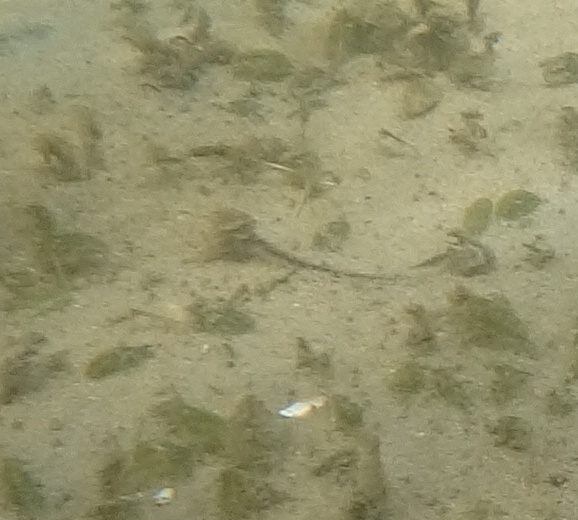} &  
    \includegraphics[width=35mm]{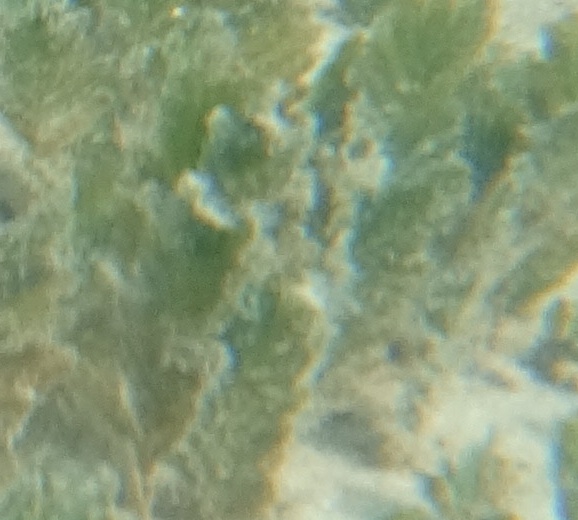} &   
    \includegraphics[width=35mm]{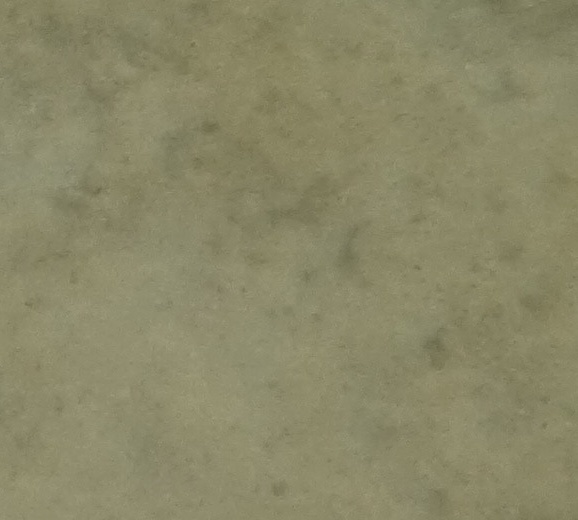}    
    \\
    \includegraphics[width=35mm]{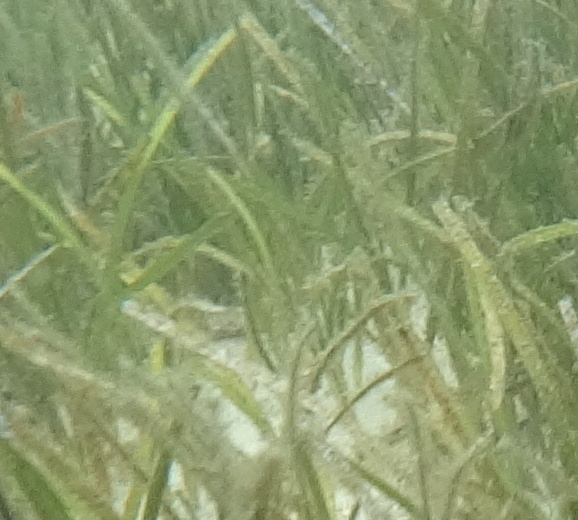} &
     \includegraphics[width=35mm]{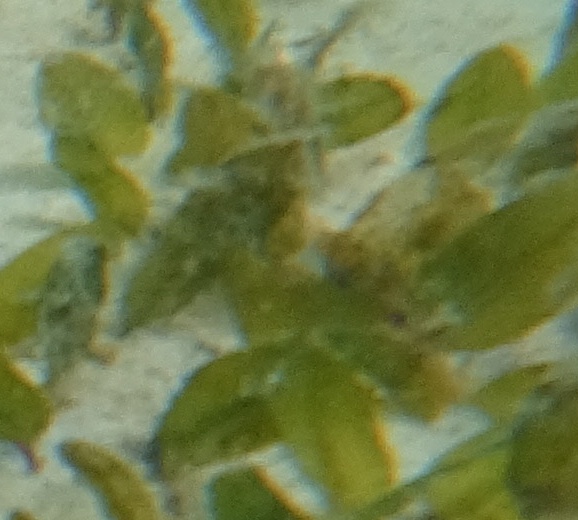} &    
    \includegraphics[width=35mm]{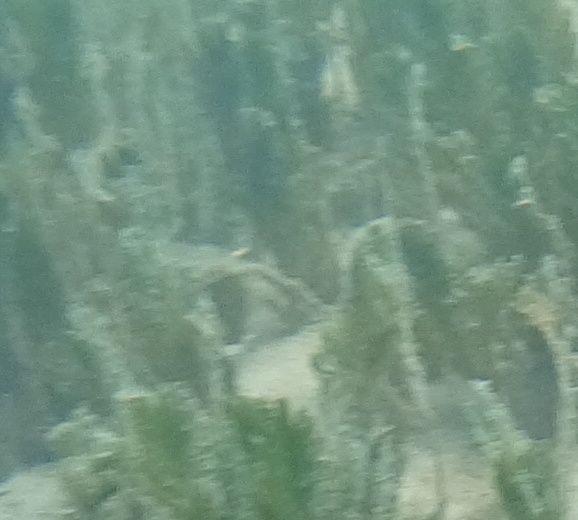} &    
    \includegraphics[width=35mm]{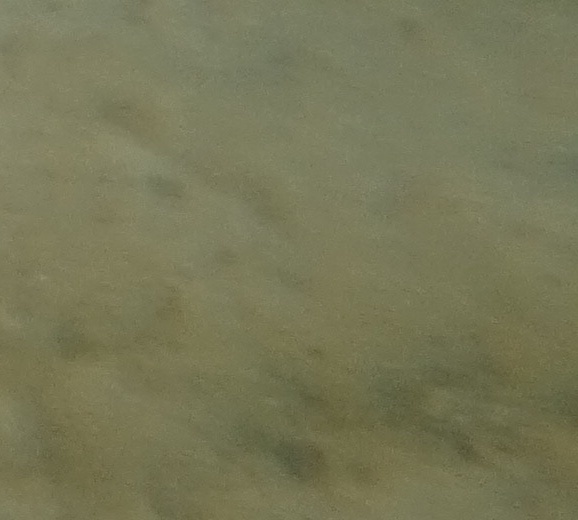}    
    \\
    \includegraphics[width=35mm]{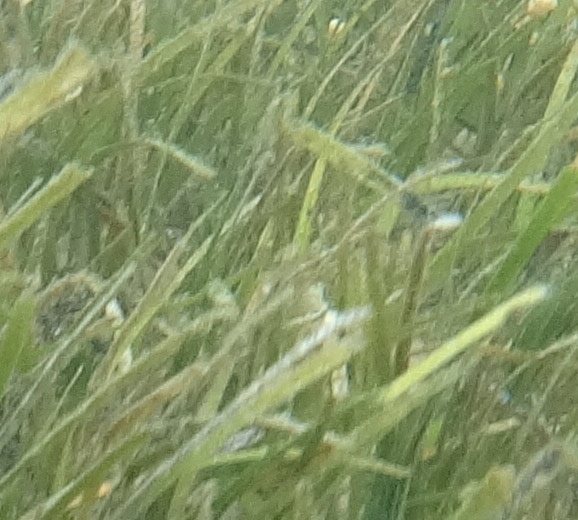} &
    \includegraphics[width=35mm]{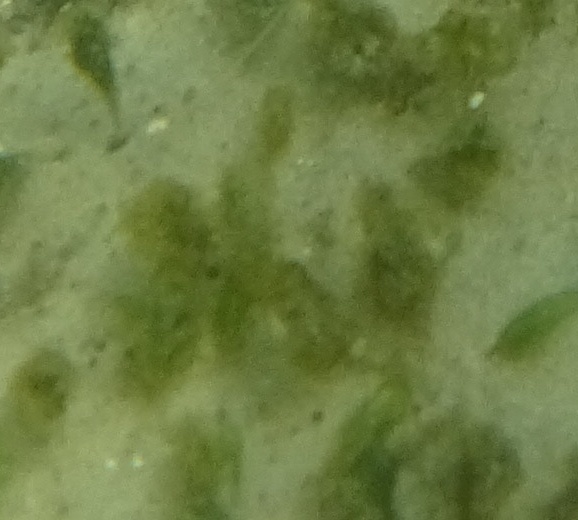} &
    \includegraphics[width=35mm]{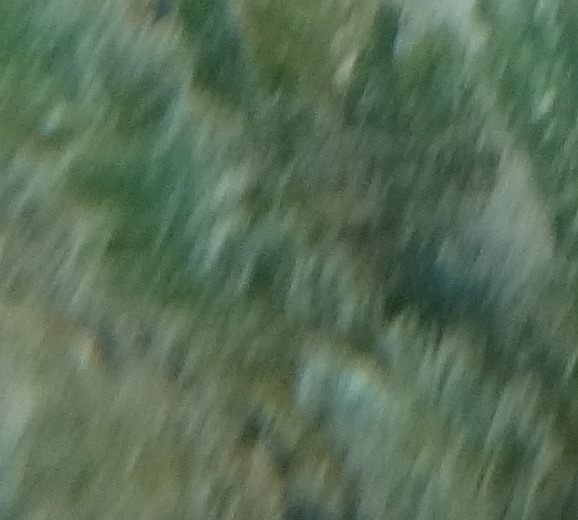} &
    \includegraphics[width=35mm]{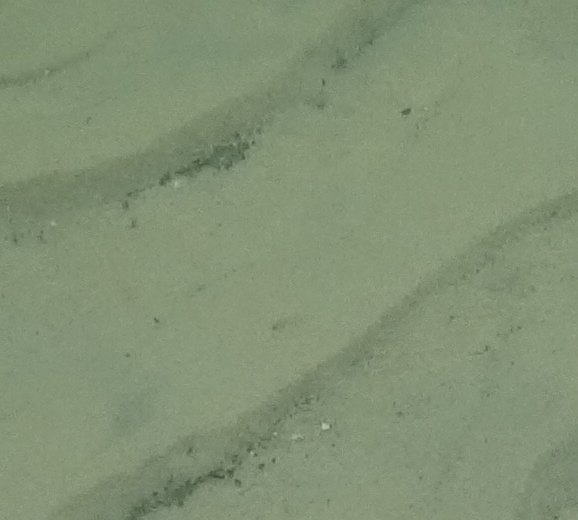}
    \\
    a) & b) & c) & d)
\end{tabular}
\captionsetup{justification=centering}
\caption{A visualisation of sample images in the DeepSeagrass dataset where images belong to the following morphotype super-classes: a) Strappy b) Rounded c) Ferny d) Background}
\label{fig:examples}
\end{figure}

Figure \ref{fig:examples} shows a selection of sample image patches from our dataset.  Each column corresponds to a different class of seagrass morphotype.

\section{Pre-Trained Models and Code}
A model trained to detect and classify seagrass on patch basis is publicly available.  The best performing model for the 4-class case i.e. Strappy morphotype, Ferny morphotype, Rounded morphotype and Background class is provided for both the 520x578 pixel patch size and also for patches of 260x289 pixels.  Additionally, we provide a pre-trained model for the 5-class case, in which we separate the Background class into sub-classes for Substrate and for the Water column.

The pre-trained models have been made available at the following link: \url{https://cloudstor.aarnet.edu.au/plus/s/nQ6JRNYvKaGqfaE}.

The models can be loaded and run using a script provided at \url{https://github.com/csiro-robotics/deepseagrass}. Instructions for setting up necessary dependencies and running the code are described in the readme file of this repository.

\section{Future Work}
In future work, we will investigate the use of domain  randomized synthetic dataset to bridge domain and species gaps \cite{ward2020scalable} \cite{ward2018deep} in the current DeepSeagrass dataset and extend it to other marine species. 

\section*{Acknowledgment}
This work was done in collaboration between CSIRO Data61, CSIRO Oceans and Atmosphere, Babel-sbf and QUT and was funded by CSIRO’s Active Integrated Matter and Machine Learning and Artificial Intelligence (MLAI) Future Science Platform.  S.R., R.M. and F.M. acknowledge continued support from the Queensland University of Technology (QUT) through the Centre for Robotics.

\bibliographystyle{IEEEtran} 
\bibliography{Bibliography}

\end{document}